\documentclass[final]{l4dc2024}

\usepackage{mathtools} 
\usepackage{wrapfig}
\usepackage{caption}
\usepackage[shortcuts]{extdash}
\usepackage{titlesec}
\usepackage{paralist}

\DeclareMathOperator*{\argmin}{arg\,min}
\DeclarePairedDelimiter{\norm}{\lVert}{\rVert} 
\newtheorem{problem}{Problem}
\newcommand{\fakepar}[1]{\vspace{1mm}\noindent\textbf{#1.}}
\newcommand{\ie}{i\/.\/e\/.,\/~}
\newcommand{\eg}{e\/.\/g\/.,\/~}
\newcommand{\cf}{cf\/.\/~}

\title[A Metric for Keypoint Detection]{Tracking Object Positions in Reinforcement Learning: \\ A Metric for Keypoint Detection}
\usepackage{times}

\coltauthor{
    \Name{Emma Cramer} \Email{emma.cramer@dsme.rwth-aachen.de}\\
    \Name{Jonas Reiher} \Email{jonas.reiher@ml.rwth-aachen.de}\\
    \Name{Sebastian Trimpe} \Email{trimpe@dsme.rwth-aachen.de}\\
    \addr Institute for Data Science in Mechanical Engineering (DSME), RWTH Aachen University \\ Dennewartstraße 27, 52068 Aachen, Germany}

\begin{document}

\maketitle

\begin{abstract}%
 Reinforcement learning (RL) for robot control typically requires a detailed representation of the environment state, including information about task-relevant objects not directly measurable. Keypoint detectors, such as spatial autoencoders (SAEs), are a common approach to extracting a low-dimensional representation from high-dimensional image data. SAEs aim at spatial features such as object positions, which are often useful representations in robotic RL. However, whether an SAE is actually able to track objects in the scene and thus yields a spatial state representation well suited for RL tasks has rarely been examined due to a lack of established metrics. In this paper, we propose to assess the performance of an SAE instance by measuring how well keypoints track ground truth objects in images. We present a computationally lightweight metric and use it to evaluate common baseline SAE architectures on image data from a simulated robot task. We find that common SAEs differ substantially in their spatial extraction capability. Furthermore, we validate that SAEs that perform well in our metric achieve superior performance when used in downstream RL. Thus, our metric is an effective and lightweight indicator of RL performance before executing expensive RL training. Building on these insights, we identify three key modifications of SAE architectures to improve tracking performance.
\end{abstract}


\section{Introduction}
\begin{figure}[tb]
    \centering
    \subfigure[SAE trained on rec. loss]{
        \label{fig:general_idea_ae}
        \includegraphics[width=57.5409836066mm]{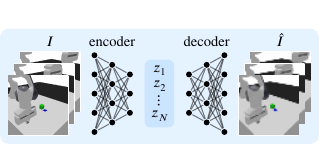} %
    }
    \hspace{5.8mm}
    \subfigure[policy trained to maximize exp.\ return]{
        \label{fig:general_idea_rl}
        \includegraphics[width=66.2786885246mm]{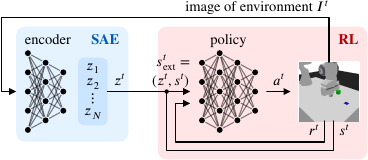} %
    }
    \caption{SAE extracts 2D positions from images via a spatial bottleneck. The SAE encoder is then integrated into an RL framework to obtain a state representation for immeasurable objects.}
    \label{fig:general_idea}
\end{figure}

In real-world control tasks like robotics, successful reinforcement learning (RL) often hinges on a thorough state representation. This necessitates including all task-relevant objects in the scene. This issue is particularly prominent in tasks involving unstructured environments or interactions with numerous objects, where defining the state space without significant prior knowledge is difficult. Image data provides a potential solution, either through direct end-to-end learning of the control signal or by first learning a low-dimensional representation of the high-dimensional data~\citep{bleher_learning_2022,levine_end--end_2016}. For practical applications, interpretability in terms of physical quantities is usually advantageous.
Spatial autoencoders (SAEs) have been effective in learning low-dimensional representations, expressed as 2D points on the image plane, referred to as keypoints. This latent representation can be used, \eg as part of the state representation of the RL agent. Figure~\ref{fig:general_idea} shows the complete learning pipeline. 

While keypoints have led to well-performing RL algorithms~\citep{kulkarni_unsupervised_2019,ghadirzadeh_deep_2017,chen_keystate_2023}, limited research has been conducted on whether SAEs effectively extract positional information of objects in the scene and, if so, how well they do this. Prior work often evaluates SAE architectures indirectly by training a downstream RL agent and evaluating the performance of the full RL pipeline~\citep{qin_keto_2020,boney_learning_2021,chen_keystate_2023}. This approach requires a lot of resources for SAE evaluation since  RL training is computationally expensive. Further, if weak RL performance is obtained, it is unclear whether RL or SAE training did not perform. Other works assess SAE performance by its training loss, which lacks insight into the physical meaningfulness of the keypoints. Some works propose a qualitative assessment by examining single keypoints on the image plane~\citep{zhang_unsupervised_2018,puang_kovis_2020}. This approach disregards the importance of consistency over trajectories. We argue that keypoints essentially serve as sensor readings and thus how well task-relevant objects are tracked over time needs to be assessed in terms of accuracy and reliability.

This paper proposes a straightforward metric for quantitatively evaluating the extraction of positional information of task-relevant objects in the latent space of SAEs. 
The proposed metric is applied to (i) train multiple base SAE architectures and compare their tracking performance, (ii) explore various improvements of these architectures, and (iii) learn a suitable RL task using keypoints from different SAEs as the state.
The evaluation reveals significant variations in the spatial extraction capability of common SAEs, emphasizing the importance of a thorough evaluation before incorporating them into RL states. Building on these insights, we propose three key modifications to substantially improve the tracking performance of common SAE architectures, resulting in, \eg a 30\,\% increase in tracking capability for the commonly used KeyNet architecture~\citep{jakab_unsupervised_2018}.
We demonstrate that the metric allows us to judge SAEs with regard to capturing physically interpretable positional features and that this metric is a good indicator of downstream RL performance. For the considered robotic manipulation task, SAE training takes approximately an order of magnitude less computational resources than RL training, making our metric an effective and lightweight indicator of RL performance before expensive RL training.
%
\section{Related Work} \label{sec:relwork}
Various approaches utilize deep neural networks (NNs) to extract state representations from images or videos~\citep{dwibedi_learning_2018,seo_reinforcement_2022}; we review the ones most related to this work.

\fakepar{Unsupervised state representation learning for RL}
Applications span from general continuous control~\citep{dwibedi_learning_2018,hafner_learning_2019} to robotic manipulation~\citep{lesort_deep_2019,rafailov_offline_2021}. Many models follow an autoencoder structure with a low-dimensional bottleneck, optimizing for input reconstruction~\citep{finn_deep_2016,yarats_improving_2021}. Some of these learn world models and recurrently capture environment dynamics in the latent representation~\citep{ha_recurrent_2018,seo_reinforcement_2022,hafner_mastering_2023}. Generally, autoencoders constrain the dimension, but not \emph{what} is captured in the latent space~\citep{yarats_improving_2021,rafailov_offline_2021}. In contrast, SAEs are constrained to capture 2D keypoint positions~\citep{finn_deep_2016}. 

\fakepar{Spatial autoencoder architectures}
SAE keypoints have successfully been used as RL state representations in robotic control~\citep{puang_kovis_2020,chen_keystate_2023}, to play Atari games~\citep{kulkarni_unsupervised_2019}, or to provide a goal description~\citep{qin_keto_2020}. 
Central to SAEs is the spatial soft-argmax layer first proposed by~\citet{levine_end--end_2016} to train an end-to-end deep visuomotor policy. \citet{finn_deep_2016} modified this approach to obtain a standalone deep SAE architecture, consisting of a convolutional encoder and fully connected decoder. \citet{jakab_unsupervised_2018} propose the KeyNet architecture, incorporating a convolutional decoder. 
These two elements, the encoder-decoder structure and the spatial soft-argmax layer, are essential to all SAEs. Many architectures build upon these blocks; incorporating feature transport mechanisms~\citep{kulkarni_unsupervised_2019}, working on error maps~\citep{gopalakrishnan_unsupervised_2020}, and reconstructing segmentation masks~\citep{puang_kovis_2020} or frame differences~\citep{sun_self-supervised_2022}. Recently, SAEs have been extended to learn 3D points~\citep{li_3d_2022,sun_bkind-3d_2023}.
While these approaches differ in the way they are trained, all aim to represent positional information. We focus our investigation on two of the most common base architectures \citep{finn_deep_2016, jakab_unsupervised_2018} as (i) they form the basis for many more complex architectures and (ii) we found that if trained correctly, they can serve as reliable feature extractors. In principle, our evaluation procedure can be applied to all of the above architectures.

\fakepar{Evaluation of SAEs}
Typically, SAEs are evaluated indirectly through compute-intensive RL or control performance~\citep{qin_keto_2020,wang_end--end_2022,boney_learning_2021} or qualitative visual assessments~\citep{zhang_unsupervised_2018,puang_kovis_2020}. 
In general, latent representations can be evaluated via reconstruction loss~\citep{finn_deep_2016}, disentanglement measurements~\citep{carbonneau_measuring_2022} or mutual information estimates~\citep{rezaabad_learning_2020}, all of which neglect the spatial 2D keypoint structure and thus do not assess the physical meaningfulness of the features.
In the computer vision domain, keypoints for image matching are evaluated by reprojecting from different views with known camera transformation~\citep{zhao_alike_2023}, which is not applicable for SAEs. \citet{jakab_unsupervised_2018} approximate labeled ground truth points as linear combinations of all keypoints. Their KeyNet SAE is evaluated with the percentage of these predicted points within a fixed distance from the labels. The same linear combination has been used by others to compute mean errors to ground truth points~\citep{zhang_unsupervised_2018,lorenz_unsupervised_2019,sun_self-supervised_2022}. \citet{kulkarni_unsupervised_2019} match keypoints to ground truth points via a min-cost assignment and compute precision and recall over trajectories. Although being quantitative, the above approaches cannot assess the quality of keypoints over trajectories and allow no statement about whether all task-relevant objects are represented. We find that both aspects are critical for use in control or RL and our method, described in Section~\ref{sec:approach}, addresses these key limitations in existing evaluation approaches.
%
\section{Problem Setting} \label{sec:problem}
We consider the general structure of an autoencoder $I \xrightarrow{h_{\mathrm{enc}, \phi}} z \xrightarrow{h_{\mathrm{dec}, \psi}} \hat{I}$, operating on an input image~$I$, which is mapped to a latent representation~$z$ via an NN encoder~$h_{\mathrm{enc}, \phi}$ and then back to a reconstructed image~$\hat{I}$ via an NN decoder~$h_{\mathrm{dec}, \psi}$ (\cf Figure~\ref{fig:general_idea_ae}).  Typically, the autoencoder is trained in an unsupervised fashion to minimize reconstruction loss $L(I,\hat{I}) = \|I-\hat{I}\|_2^2$
while restricting the dimension of the latent space with a low dimensional bottleneck.
Here, we are particularly interested in spatial autoencoders (SAEs)~\citep{finn_deep_2016}, which aim to represent 2D positions of objects in an image as latent variables $z$.  For this, the last layer of the encoder $h_{\mathrm{enc}, \phi}$ with $N$ outputs is chosen as a soft-argmax layer according to~\citet{finn_deep_2016}. This layer ensures that the latent space can be interpreted as $N$ keypoints in the image plane with $z = (z_1, z_2, \dots, z_N) \in \mathbb{R}^{2 \times N}$. For this, the feature maps $M \in \mathbb{R}^{H\times W \times N}$ of the last convolutional encoder layer are passed through a channel-wise softmax layer $s_{hwn}=\exp(m_{hwn} / \alpha) / \sum_{h^{\prime},w^{\prime}} \exp(m_{h^{\prime} w^{\prime} n} / \alpha)$, where $\alpha$ is a learned temperature parameter and $h$, $w$, and $n$ are indices along the height, width, and depth dimensions of $M$. Then the $n$-th 2D point of maximum activation is computed as $z_n=(\sum_{h,w} h\cdot s_{hwn}, \sum_{h,w} w \cdot s_{hwn})$. 

We consider a setup with $K$ rigid objects that shall be tracked.  Let the ground truth position of the $k$-th object (\eg its center of mass) in the 2D image space be given by $x_k \in \mathbb{R}^2$, and the positions of all objects collectively by $x = (x_1, \dots, x_K) \in \mathbb{R}^{2 \times K}$. 
An ideal SAE should track $x$ with its latent representation $z$ in \emph{some sense}. However, how to evaluate the tracking performance is unclear, and proposing a method that quantifies this is our main objective:
\begin{problem}
\label{probl:sae_metric}
We seek to quantify how well the keypoints~$z$ represent the ground truth objects~$x$.
\end{problem}
SAEs are often used as feature extractors for RL tasks, where keypoints $z$ are then part of the state representation.
In RL, an agent learns to optimize an objective through interaction with an environment~\citep{sutton_reinforcement_1998}. The environment is represented as a discounted Markov decision process (MDP) defined by the tuple $(\mathcal{S}, \mathcal{A}, p, r, \rho_0, \gamma)$, with state space $\mathcal{S}$, action space $\mathcal{A}$, the typically unknown transition probability distribution $p: \mathcal{S} \times \mathcal{A} \times \mathcal{S} \rightarrow \mathbb{R}$, the reward function $r: \mathcal{S} \times \mathcal{A} \rightarrow \mathbb{R}$, the distribution of the initial state $\rho_0(s_0): \mathcal{S} \rightarrow \mathbb{R}$, and the discount rate $\gamma \in (0, 1)$. 
A policy $\pi: \mathcal{S} \times \mathcal{A} \rightarrow \mathbb{R}$ selects an action with a certain probability for a given state. 
The agent interacts with the MDP to collect episodes $\tau = (s^0, a^0, r^1, s^1, \dots, s^T)$, which are sequences of states, actions, and rewards over time steps $t=0, \dots, T$. The usual objective in RL is to find the policy $\pi$ that maximizes the expected return $J(\pi) = \mathbb{E}_{\tau} [\,\sum_{t=1}^{T} \gamma^t r^t\,]$, where the expectation is over trajectories $\tau$ under the policy $\pi$. 
The general understanding in literature~\citep{finn_deep_2016,ghadirzadeh_deep_2017,kulkarni_unsupervised_2019,wang_end--end_2022} is that well-tracking SAEs will yield better RL performance, such as higher expected return $J(\pi)$ or episode success rates.
We investigate whether this holds true for the metric proposed for Problem~\ref{probl:sae_metric}:
\begin{problem}
    Is SAE performance (according to Problem~\ref{probl:sae_metric}) an indicator for RL performance?
    \label{probl:rl_performance}
\end{problem}
If this hypothesis holds true, SAE performance can be evaluated before actual RL training, usually at significantly lower computational cost.
%
\section{A Metric to Evaluate Keypoints}
\label{sec:approach}
In this section, we propose a metric to quantify the tracking performance of an SAE, addressing Problem~\ref{probl:sae_metric}. In Section~\ref{sec:sae_eval}, we then use the metric for RL to evaluate Problem~\ref{probl:rl_performance}.  

As RL makes decisions sequentially over time, we are interested in tracking performance over multiple time steps.  Therefore, we denote by $x_k^t \in \mathbb{R}^2$ the ground truth position of the $k$-th object, and by $x^t \in \mathbb{R}^{2 \times K}$ the positions of all $K$ objects collectively at time $t$.
Furthermore, we denote the trajectory of these objects over time steps $t=0, \dots, T-1$ by $x^{\tau} = (x^{0}, x^{1}, \dots, x^{T-1}) \in \mathbb{R}^{2 \times K \times T}$. 
We use analogous notation for the $N$ latent keypoints; that is, $z^{\tau} = (z^{0}, z^{1}, \dots, z^{T-1}) \in \mathbb{R}^{2 \times N \times T}$ denotes the trajectory of keypoints.

Given this notion of trajectories, Problem~\ref{probl:sae_metric} translates to measuring how well the keypoint trajectory~$z^{\tau}$ follows the ground truth trajectory~$x^{\tau}$ for a given instance of an SAE.
A naive approach would be to directly compute the Euclidean distance between point-pairs along these trajectories. However, this will not yield satisfactory results.  The keypoints are learned in an unsupervised fashion, which provides no guarantee about which part of an object is tracked. 
As points on a rigid object have a fixed relation to each other, it is reasonable to assume that, for downstream RL training, any point on the object is an equally suitable representation.
For example, if a ground truth point and a keypoint are on the same object at a constant offset, this offset would accumulate to a tracking error when naively taking the difference between the two points. Thus, we need to account for offsets by an appropriate transformation.
Finally, the SAE extracts many keypoints (usually $N > K$) and the association of keypoints to ground truth points is unknown.
Taking these together, an evaluation protocol of keypoints will thus require (i) accounting for the offset between any point on the object and ground truth, (ii) associating keypoints with ground truth points, and (iii) developing a quantitative measure to evaluate the capability of tracking all relevant ground truth points.

\begin{wrapfigure}{r}{60mm}
    \centering
    \raisebox{0pt}[\dimexpr\height-1.0\baselineskip\relax]{%
    \includegraphics[width=18.3mm]{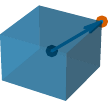}
    \hfill
    \includegraphics[width=37mm]{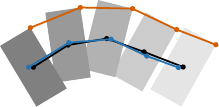}
    }%
    \caption[format=plain]{We consider keypoints (red) to be equally informative about object positions as the ground truth CM (black). Motion of both points results in a varying offset in the image plane. We evaluate with transformed keypoints (blue) minimizing the offset.}
    \vspace{-1\baselineskip}
    \label{fig:transformation}
\end{wrapfigure}

\fakepar{Transformation}
Keypoints are coordinates in the 2D image space, which are supposed to track objects in 3D space. Often, the center of mass (CM) is taken as the ideal point to represent the 3D position of an object in the world frame. However, for the downstream RL task, the keypoints do not have to track the CM, but any fixed point on the object, \ie the point's offset from the CM should be constant in the object's 3D frame of reference (\cf Figure~\ref{fig:transformation}).
If the keypoints were to track the CM, keypoints and ground truth points would coincide in image space. Due to the 3D offset, we also observe an offset in 2D-image space (\cf Figure~\ref{fig:transformation}). 
This 2D offset is generally unknown; it depends on the unknown 3D offset, object position, orientation, and camera view.
Even if the keypoints were to track a point on an object perfectly, this offset would falsely suggest a tracking error in 2D.
Instead of capturing the full geometry of the problem, which requires additional problem insight, we propose a lightweight approach that eliminates the main offsets between keypoints and ground truth.  We consider a time-invariant affine transformation of keypoints $\hat{z} = A z + b$, where $A \in \mathbb{R}^{2 \times 2}$ and $b \in \mathbb{R}^2$ are fit via ordinary least squares on a held-out test set, containing random time steps from trajectories unseen in SAE training. This transformation can account for the scaling and translation of a keypoint trajectory. 
We note that even with a time-invariant 3D offset, the 2D offset can be time-variant due to the object's motion; the time invariance thus represents an approximation.
Still, we find that this transformation is easy to compute, requires no additional information about the ground truth objects, and works well in practice (\cf Section~\ref{sec:evaluation}).

\fakepar{Association and tracking error}
\label{sec:tracking_error}
Consider the trajectories~$x^{\tau}$, $\hat{z}^{\tau}$ of $K$ ground truth objects and $N$ transformed keypoints.
We propose an error metric between the trajectory of one ground truth object $x^{\tau}_n$ and the transformed trajectory of one keypoint $\hat{z}^{\tau}_n$. We define the tracking error~$e_{n,k}$ between any two trajectories $n$ and $k$ as 
\begin{equation}
    e_{n,k} = \sum_{t=1}^{T} \norm{\hat{z}^t_n - x^t_k}_2^2.
    \label{eq:error}
\end{equation}
The error~$e_{n,k}$ is a measure of how well a specific keypoint tracks a ground truth object over time. 
Using the tracking error, we determine the index of the keypoint $z_{n^*_k}$ that best tracks object $x_k$ as $n^*_k = \argmin_{n}  e_{n,k}$.
Once we assigned the most suitable keypoint for each ground truth object, we give the tracking error of the associated keypoint as $e_{n^*_k,k}$. For our evaluation, we always consider the tracking error of the best keypoint. The lower this tracking error, the better the ground truth point~$x_{k}$ is represented by the keypoint~$z_{n^*_k}$. 
The error measure enables a comparison of different SAE architectures and individual training runs of the same architecture broken down into objects.

For the later evaluation of SAEs, we now define indicators for an SAE's overall tracking performance.
We classify an object~$x_k$ as correctly tracked if the tracking error of the most suitable keypoint~$z_{n^*_k}$ is below an application-specific threshold~$\mu_k>0$.
The index set~$\mathcal{X}_c$ of all correctly tracked objects is given by $\mathcal{X}_c = \{k : e_{n^*_k,k} \leq \mu_k\}$.
We then define the tracking capability~$\mathrm{TC}$ of one trained SAE as the percentage of tracked ground truth objects, \ie
\begin{equation}
    \mathrm{TC} = \left| \mathcal{X}_c \right| / K.
    \label{eq:tracking_capability}
\end{equation}
An ideal tracking capability of $\mathrm{\mathrm{TC}}=1$ means that for this SAE, the position of all ground truth objects is correctly encoded in the latent space.

A quantitative evaluation should consider the distribution of the tracking error and the tracking capability over multiple training runs.
We look at the mean, median, and the variance of the tracking error over runs.
Similarly, we evaluate the mean tracking capability~$\overline{\mathrm{TC}}$ over multiple runs. 
Intuitively, $\overline{\mathrm{TC}}$ gives the mean percentage of all ground truth objects captured by keypoints. An SAE with a high mean tracking capability is a reliable feature extractor for RL scenarios. For individual ground truth objects, we denote as $\overline{\mathrm{TC}}_k$ the mean tracking capability for object~$k$.
%
%
%
\section{Evaluation}
\label{sec:evaluation}

\begin{wrapfigure}{r}{60mm}
    \centering
    \vspace{-2.7\baselineskip}
    \includegraphics[width=18mm]{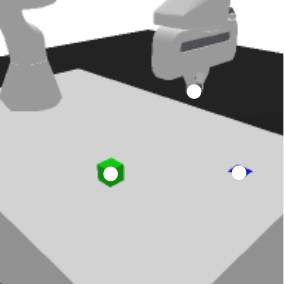}
    \hfill
    \includegraphics[width=18mm]{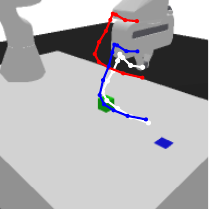}
    \hfill
    \includegraphics[width=18mm]{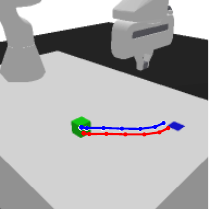}
    \caption{The PandaPush-v3 task with three object positions $x_k$ marked (left). Selected ground truth, keypoint, and transformed keypoint trajectories are shown in white, red, and blue for the end effector (middle) and cube (left).}
    \vspace{-1\baselineskip}
    \label{fig:pandapush}
\end{wrapfigure}

We first use our proposed metrics \eqref{eq:error} and \eqref{eq:tracking_capability} to evaluate the tracking performance of base SAE architectures commonly used in RL and propose architecture modifications to improve tracking.  We then investigate how tracking performance links to performance in a downstream RL task. 
The empirical results reveal the following main insights:
\begin{compactenum}
    \item The proposed metric is able to quantify the tracking performance of SAEs.
    \item The combination of the best baseline SAE with our proposed modification yields $\overline{\mathrm{TC}} = 0.99$, and it can thus be considered a reliable and precise spatial feature extractor.
\end{compactenum}

\begin{compactenum}
    \setcounter{enumi}{2}
    \item Our proposed metric for SAE tracking performance is indicative of the performance of RL; that is, the architecture with best SAE metric also achieves best asymptotic return. 
    \item The best-found architecture in terms of SAE tracking achieves an RL return comparable to training with ground truth points.
\end{compactenum}
%
\subsection{SAE Evaluation}
\label{sec:sae_eval}

We demonstrate the suitability of the tracking error and tracking capability introduced in Section~\ref{sec:approach} to evaluate the performance of SAEs. We provide visualizations of our quantitative results at \href{https://youtu.be/8KqFXQiWa9w}{youtu.be/8KqFXQiWa9w}.

\fakepar{SAE experiment setup}
\label{sec:experiment_setup}
We use the PandaPush-v3 environment from the panda-gym~\citep{gallouedec_panda-gym_2021} simulation. The robot's task is to push a cube toward a target (\cf Figure ~\ref{fig:pandapush}). We identify three task-relevant objects in this environment, (i) the green \emph{cube} to be moved, (ii) the blue square indicating the \emph{target}, and (iii) the tip of the \emph{end effector}. 
The different sizes and motion behavior of the objects make them a suitable selection to evaluate the tracking performance.
We consider three standard SAE architectures and our own combination of modifications: 
\begin{compactenum}
    \item[\textbf{Basic}:] We design the Basic architecture to be a simple and efficient SAE baseline incorporating the key components that all SAEs typically share. The CNN encoder has six convolutional layers and max-pooling operations in between. The decoder uses KeyNet's Gaussian kernel maps, followed by three convolutional layers. We look at two versions of this SAE with $N=16$ (Basic) and $N=32$ keypoints (Basic-kp32).
    \item[\textbf{DSAE}~\citep{finn_deep_2016}:] DSAE introduced the spatial soft-argmax bottleneck, still used in many other architectures~\citep{zhang_unsupervised_2018,cabi_scaling_2019,gopalakrishnan_unsupervised_2020,puang_kovis_2020,boney_learning_2021}.
    This was the first SAE to be successfully used for RL training. $N=16$ keypoints are captured between a CNN encoder and fully connected decoder.
    \item[\textbf{KeyNet}~\citep{jakab_unsupervised_2018}:] KeyNet is a widely used and built upon SAE architecture~\citep{kulkarni_unsupervised_2019,minderer_unsupervised_2019,gopalakrishnan_unsupervised_2020}, consisting of a CNN encoder and decoder with $N=30$ keypoints. Input to the decoder are $N$ feature maps with isotropic Gaussian kernels at the corresponding keypoint locations. 
    \item[\textbf{Vel-std-bg modifications}:] We propose a set of modifications to the above architectures, combining ideas from existing works and new ones. Analogously to DSAE, we add a velocity loss term to the reconstruction loss with a weighting factor~$\beta$. By penalizing a change of keypoint velocities in subsequent frame pairs, the velocity loss encourages temporal consistency. KeyNet uses Gaussian heatmaps as input to the first CNN decoder layer. We propose making the standard deviation $\sigma$ of these heatmaps trainable. This enables the decoder to control the radius of influence of a keypoint. Finally, we add a bias with the dimensions of the target image to the decoder's output, giving the decoder a straightforward way to reconstruct a stationary background and allowing time-varying keypoints to focus on moving objects. For the modified architectures, we call the combinations of the KeyNet or Basic architecture combined with our proposed modifications KeyNet-vel-std-bg and Basic-vel-std-bg, respectively.
\end{compactenum}
While many more architectures exist in literature (\cf Section \ref{sec:relwork}), we deliberately choose baseline architectures maintaining the usual autoencoder setup without auxiliary networks such as adversaries or feature transport mechanisms. We choose modifications which we believe to be beneficial for the main goal of SAEs, spatial tracking of keypoints over time.
For SAE tracking evaluation, we conduct 24 training runs with different random seeds. The tracking thresholds need to be chosen heuristically. Here we choose $\mu_\mathrm{cube} = \mu_\mathrm{target} = 0.015$ and $\mu_\mathrm{eef} = 0.1$. Intuitively larger objects result in a larger tracking error, due to the possible offset to the center of mass. 
We find a good heuristic to be related to the SAE reconstruction. Objects appear in the reconstruction when the tracking error falls below $\mu_k$.
Additional modifications, an ablation study, and all experimental details such as hyperparameters can be found in the appendix.

\fakepar{Evaluating accuracy via the tracking error}
\begin{figure}[tb]
    \centering
    \includegraphics[width=130mm]{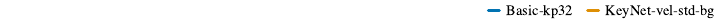} 
    \includegraphics[width=130mm]{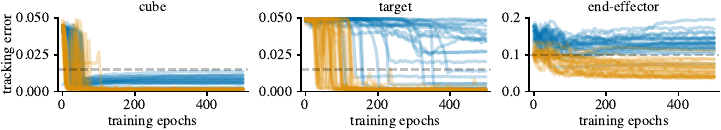} 
    \caption{Basic-kp32 and KeyNet-vel-std-bg tracking errors $e_{n^*_k,k}$ for $K=3$ objects over epochs.}
    \label{fig:tracking_error_time}
\end{figure}
First, we study the tracking error of individual runs during training. We observe a sudden drop in tracking error whenever the SAE has learned to track an object. 
To understand this behavior, we look at the tracking error over episodes for an SAE with medium performance, the Basic-kp32, in Figure~\ref{fig:tracking_error_time}.
All runs for Basic-kp32 show the drop in tracking error for the cube, which is the easiest to track. For the target, which is slightly harder to track due to its smaller size and rare movement, only a few runs show the expected drop below our threshold, resulting in a correctly tracked target. 
Instead of a sharp drop, the tracking error for the end-effector shows a shallow decrease over training epochs. We interpret this observation as follows: The end-effector occupies considerably more pixels in the image than cube and target. Thus, the reconstruction first focuses on these areas, resulting in early vague tracking and reconstruction. However, tracking a point on the end-effector consistently is achieved only by a few runs.
Looking at the tracking error of KeyNet-vel-std-bg, Figure~\ref{fig:tracking_error_time} shows a distinct drop below the threshold for the cube and target. Even for the end-effector, the tracking error consistently falls below the threshold, indicating successful tracking. 
We find that the tracking error is useful in examining exactly how accurate a trained SAE architecture instance can track individual objects.

\fakepar{Evaluating reliability via the tracking error}
\begin{figure}[tb]
    \centering
    \includegraphics[width=130mm]{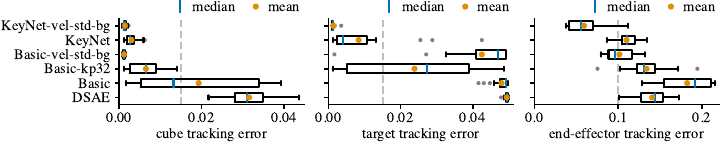} 
    \caption{Box plots of the tracking error $e_{n^*_k,k}$ for $K=3$ ground truth objects.}
    \label{fig:sae_boxplots}
\end{figure}
Our results indicate differing tracking performance for random seeds within the same architecture, showing that the SAE architectures need to be evaluated over multiple training runs.
The tracking error's distribution over 24 training runs is illustrated in Figure~\ref{fig:sae_boxplots}. We remark that the tracking error varies among (i) architectures, (ii) random seeds, and (iii) objects. Among the standard architectures, KeyNet attains the lowest mean tracking error and smallest variance, indicative of good overall tracking performance. For DSAE and Basic, larger tracking errors with greater variance are observed, marking them less reliable. The KeyNet-vel-std-bg architecture shows lowest mean tracking error and variance for all three objects.
We identify the criteria for well-performing architectures as low mean tracking error and small variance over runs.

\fakepar{Evaluating overall performance via the tracking capability}
\begin{figure}[tb]
    \centering
    \includegraphics[width=130mm]{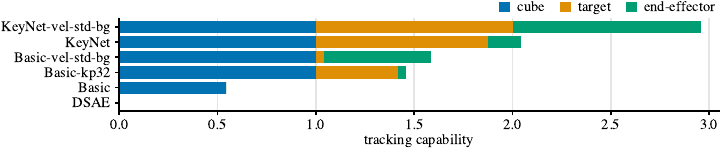} 
    \caption{Tracking capabilities $\overline{\mathrm{TC}_k}$ for $k=3$ ground truth objects of the baseline architectures.}
    \label{fig:tracking_capabilities}
\end{figure}
Figure~\ref{fig:tracking_capabilities} shows the sum of mean object tracking capabilities~$\overline{\mathrm{TC}}_k$ over architectures, further demonstrating the varying tracking performance across SAE architectures. Examining the tracking capability with regard to the individual ground truth objects, we further substantiate our hypothesis that the target and end-effector are more difficult to track than the cube. The tracking performance of all base architectures has potential for improvement as none is close to the theoretical maximum of 3.0. The combination vel-std-bg yields consistent improvement in tracking capability. 
KeyNet already tracks cube and target well and has a $\overline{\mathrm{TC}}=0.681$. KeyNet-vel-std-bg has a near-perfect mean tracking capability of $\overline{\mathrm{TC}}=0.986$. The biggest change can be seen in end effector tracking, which improved from $\overline{\mathrm{TC}}_\mathrm{eef}=0.167$ to $0.958$.
We see that the tracking capability is a compact description of how well task-relevant objects are tracked. This information is critical for downstream control and RL tasks.

Combining the insights from the tracking error and tracking capability answers Problem \ref{probl:sae_metric}.

\subsection{RL Evaluation}
We run RL experiments with SAE architectures selected by their tracking performance and find that this is a good indicator of downstream RL performance.

\fakepar{RL experiment setup}
For RL experiments with SAEs as state, we randomly sample 5 trained SAEs per architecture and conduct 2 randomly seeded RL training runs with each of them, yielding a total 10 runs per SAE architecture. We use the SAC~\citep{haarnoja_soft_2018} implementation from stable-baselines3~\citep{raffin_stable-baselines3_2021}. Hyperparameters are listed in appendix~\ref{sec:implementation_details}.

We consider two types of state representation for RL with SAE-encoded keypoints: (i) latent keypoints as state $s^t = z^t$, (ii) latent keypoints combined with robot 3D position~$o_\mathrm{eef}$ and velocity~$\dot{o}_\mathrm{eef}$, giving $s^t_\mathrm{ext} = (z^t, o_\mathrm{eef}, \dot{o}_\mathrm{eef})$. The second scenario is relevant since end-effector position and velocity are often available as robot state measurements.
As additional benchmarks, we include state representations with ground truth points $x^t$, which are usually not available in practice, obtaining $s^t = x^t$ and $s^t_\mathrm{ext} = (x^t, o_\mathrm{eef}, \dot{o}_\mathrm{eef})$. Finally, we compare to RL runs with the full 3D simulation state, including positions, velocity, and orientation of cube and target.
Actions consist of 3D displacements~$a^t = (\Delta o_\mathrm{eef})^t$ of the end effector at every time step. We use a sparse reward with~$r_t = -1$ and $r_{T}=0$ on episode success.
Following~\citep{agarwal_deep_2021}, we report interquartile mean (IQM) success rates with bootstrapped 95\,\% percentile confidence intervals.

\begin{figure}[tb]
    \centering
    \includegraphics[width=130mm]{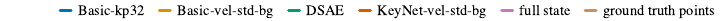} 
    \subfigure[state $s^t$]{
        \label{fig:rl_fps_only}
        \includegraphics[width=63.7mm]{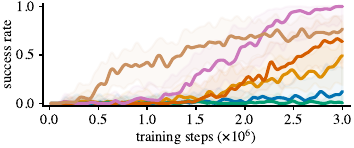} 
    }
    \hspace{2.6mm} 
    \subfigure[state $s^t_\mathrm{ext}$]{
        \label{fig:rl_fps_ee}
        \includegraphics[width=63.7mm]{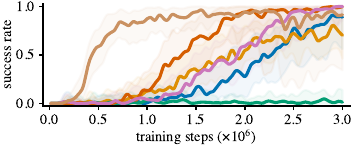} 
    }%
    \caption{Success rate for different states: keypoints only in \ref{fig:rl_fps_only} and extended state in \ref{fig:rl_fps_ee}.}
\end{figure}

\fakepar{Reinforcement learning with keypoints}
\label{sec:RL_eval}
Figure~\ref{fig:rl_fps_only} shows the RL performance with state representation $s^t$. We observe varying success rates depending on the SAE architecture and the gradations in RL performance follow the order of SAEs by tracking capability, as seen in Figure~\ref{fig:tracking_capabilities}. The DSAE architecture shows no RL progress. Although both Basic-vel-std-bg and Basic-kp32  have similar total tracking capabilities (\cf Fig.~\ref{fig:tracking_capabilities}), the former performs better on the RL task. This is due to its ability to track the end-effector reasonably well, while Basic-kp32 tracks the target instead. End-effector tracking is critical, as moving the cube is otherwise impossible. 
The best-tracking KeyNet-vel-std-bg dominates the RL with learned keypoints. Still this architecture does not reach the full-state performance. This is to be expected since the representation is limited to 2D space and lacks velocity information. The runs using 2D ground truth points, mimicking a perfect SAE, learn significantly earlier than KeyNet-vel-std-bg, but only achieve a slightly higher final success rate.

Figure~\ref{fig:rl_fps_ee} shows the RL runs using state $s^t_\mathrm{ext}$, \ie including the end-effector's 3D position and velocity in addition to keypoints. RL performance with KeyNet-vel-std-bg and the full state show comparable final success rates close to $1.0$, indicating that 2D keypoints are a useful state representation. KeyNet-vel-std-bg additionally learns notably faster. We presume that the reduced 2D representations of target and cube positions accelerate RL training.
As expected with DSAE, tracking neither target nor cube, learning progress is impossible. Compared to the first experiment setup, Basic-kp32 and Basic-vel-std-bg switch positions in final RL performance. Although Basic-vel-std-bg improves with the more precise 3D end-effector position, it is still unable to track the target and, therefore, limited in performance. For Basic-kp32, the missing end-effector tracking is now compensated with ground truth 3D information. Using its notable target tracking capability, it achieves better final performance. Initially, Basic-vel-std-bg learns faster, supporting the assumption that 2D representations can accelerate RL training. These kinds of insights are facilitated by the tracking capability and would not have been possible via traditional SAE evaluation. The IQM success rates for runs with ground truth instead of keypoints show faster learning but do not quite reach the maximum of full state and KeyNet-vel-std-bg.

Answering Problem \ref{probl:rl_performance}, we find a link between SAE tracking capability, including the tracking capability for individual objects, and downstream RL performance. 
%
%
\section{Conclusion}
\label{sec:conclusion}
We propose a metric to evaluate SAE performance with respect to task-relevant objects. By means of this metric, we show that well-performing SAE architecture actually track positions of task-relevant objects. We find notable performance differences in SAE architectures and identify three components that reliably improve performance, leading to almost perfect object tracking. We show that SAE tracking performance is indicative of downstream RL performance for a representative robotic manipulation task. This allows identifying suitable SAEs after comparatively lightweight SAE pretraining and before computationally expensive RL training. In addition, troubleshooting is greatly facilitated by the ability to evaluate the performance of an SAE as a key component of the RL pipeline.
We observe that an RL agent using keypoints as part of its state achieves RL performance comparable to an agent with full simulation state. Thus, we consider keypoints a suitable state representation for robotic RL.
We have demonstrated that this straightforward metric is effective in evaluating SAE architectures. The metric can be used to analyze any 2D keypoint extractor and is not restricted to SAEs. Investigating alternative keypoint extractors and extensions to 3D keypoints is thus a promising avenue for future research.
The code to reproduce all results is available at \href{https://github.com/Data-Science-in-Mechanical-Engineering/SAE-RL}{github.com/Data-Science-in-Mechanical-Engineering/SAE-RL} and can be used to inform future research.

\acks{We thank Paul Brunzema and Bernd Frauenknecht for their helpful comments. We also thank Robin Kupper for his contributions in the early stages of this research. This work was partially funded by the “Demonstrations- und Transfernetzwerk KI in der Produktion (ProKI-Netz)” initiative, funded by the German Federal Ministry of Education and Research (BMBF, grant number 02P22A010). Computations were performed with computing resources granted by RWTH Aachen University under project rwth1385.}

\bibliography{main}
%
%
\newpage
\appendix
\section{Implementation Details}
\label{sec:implementation_details}
The following section gives more details about our implementation and training framework.

\subsection{SAE Experiment Setup}
For SAE tracking evaluation, we conduct 24 training runs with different random seeds for each of the considered model architectures.

\fakepar{Training}
Hyperparameters are kept fixed and are listed in Table~\ref{tab:sae_hyperparameters}. During SAE training, we add normally distributed noise with $\sigma = 0.001$ to the input images for regularization. Images contain floating point RGB values in the range $[0, 1]$.
\begin{table}
    \centering
    \caption{Hyperparameters for SAE training.}
    \begin{tabular}{| l | l |}
        \hline
        SAE Hyperparameter &  Value\\
        \hline
        number of epochs & $500$ \\
        batch size & 32 \\
        learning rate & 0.001 \\
        optimizer & Adam~\citep{kingma_adam:_2014} \\
        input shape & $256 \times 256 \times 3$ \\
        target shape & $64 \times 64 \times 3$ \\
        latent spatial dimensions & $[-1, 1] \times [-1, 1]$ \\
        \hline
    \end{tabular}
    \label{tab:sae_hyperparameters}
\end{table}

\fakepar{Datasets}
The training, validation, and testing datasets consist of $5\,000$, $2\,500$, and $2\,500$ images, respectively. To keep all three datasets well-separated, we initially collect image sequences of 10 frames each and perform data splitting on these sequences. To capture sufficient variation in the randomly initialized object and target positions, episode lengths during image collection are limited to $20$ frames. Images are collected using a smoothed random policy, starting with a random initial action in $[-10, 10] \times [-10, 10] \times [-10, 10]$ and sampling the next action from a Gaussian normal distribution with standard deviation $\sigma = 1.0$ centered on the previous action.

\fakepar{Evaluation}
The tracking thresholds used for computing the tracking capability $\mathrm{TC}$ are chosen to be $\mu_\mathrm{cube} = 0.015$, $\mu_\mathrm{target} = 0.015$, and $\mu_\mathrm{eef} = 0.1$. Tracking errors are evaluated in the latent spatial dimensions of size $[-1, 1] \times [-1, 1]$. Where point estimates are are reported, they refer to values at the end of training. Figures showing the object-wise tracking errors over time are slightly smoothed by applying a Gaussian filter with $\sigma = 2.5$ steps.

\subsection{SAE Architectures}
Existing and new SAE architectures were implemented by us in PyTorch~\citep{paszke_pytorch_2019}. All details of this implementation can be found in the accompanying source code.

\fakepar{Architectures from literature} Our DSAE implementation follows the layout outlined by~\citet{finn_deep_2016}. We keep all mentioned hyperparameters, including a CNN encoder with three convolutional layers, the number of keypoints~$N = 16$ and a single fully connected layer as decoder.
Our KeyNet implementation follows the implementation from~\citet{jakab_unsupervised_2018}. In their paper, the goal is to shift an object in one image with information from another image. For this, a second image is appended to the decoder input. We neglect this since we are assuming that the scene does not change. Where applicable, we keep all mentioned hyperparameters. We choose the number of keypoints to be~$N = 30$, as primarily used in their experiments. We fix the number of convolutional blocks in the encoder as $4$ and in the decoder as $3$.

\fakepar{Basic} 
Our Basic SAE architecture is a CNN similar to KeyNet~\citep{jakab_unsupervised_2018}. The encoder consists of three convolutional blocks. Each block consists of a 2D convolution followed by a 2D batch-normalization layer and another 2D convolution followed by a 2D max-pooling operation with kernel size and stride $2$ and a 2D batch-normalization layer again. The kernel size for every convolution is $3$. The number of output channels for both convolutions in a block is identical with $32$ in the first, $64$ in the second, and $128$ output channels in the third block. The blocks are followed by a 2D $1 \times 1$-convolution to aggregate into $16$ feature maps. A spatial soft arg-max layer extracts $N = 16$ 2D keypoints from these. Similarly to KeyNet, the decoder begins with 2D heatmaps with Gaussian kernels with $\sigma = 0.1$ at the keypoint location for each of the $N = 16$ feature maps. These maps are created at the target resolution of $64 \times 64$. Three convolutional layers follow, each with a kernel size of $3$ and $64$, $32$, and $16$ output channels, respectively. Each convolution is followed by a 2D batch-normalization layer. A final 2D $1 \times 1$-convolution again aggregates the feature maps into $3$ RGB image channels.

\fakepar{Modifications}
We find three modifications to SAE architectures that improve tracking:
\begin{compactenum}
    \item \textbf{Velocity loss term} (-vel). In~\citet{finn_deep_2016}, an additional loss term~$g_\mathrm{slow}$ is added to the reconstruction loss with a weighting factor~$\beta$. By penalizing a change of keypoint velocities in subsequent frame pairs, the velocity loss encourages the detection of temporally consistent keypoints.
    For the velocity loss term, we choose a weighting factor $\beta = 0.1$ while the reconstruction loss remains weighted with factor $1.0$ and is here computed as the mean of the three samples passed through the encoder concurrently.
    \item \textbf{Trainable Gaussian standard deviation} (-std). We follow~\citet{jakab_unsupervised_2018} and use Gaussian heatmaps as input to the first CNN decoder layer. We propose making the standard deviation $\sigma$ trainable. This enables the decoder to control a keypoint's radius of influence.
    For the trainable Gaussian standard deviation, we use an initial value of $\sigma = 0.1$, equal to the fixed standard deviation in the unmodified case.
    \item \textbf{Background bias layer} (-bg). We add a bias with the dimensions of the target image to the decoder's output, giving the decoder a straightforward way to reconstruct a stationary background. This usually happens within the first epochs and time-varying keypoints then focus on moving objects. The background bias layer is initialized to zeros.
\end{compactenum}

\subsection{RL Experiment Setup}
For RL experiments with SAEs as state representation extractors, we randomly sample 5 trained SAEs of each architecture and conduct 2 randomly seeded RL training runs with each of them, yielding a total of 10 runs per SAE architecture.

\fakepar{Training}
As an RL algorithm, we choose the SAC~\citep{haarnoja_soft_2018} implementation from stable-baselines3~\citep{raffin_stable-baselines3_2021}. Except for the values in Table~\ref{tab:rl_hyperparameters}, hyperparameters are kept at default values.
\begin{table}
    \centering
    \caption{Hyperparameters for RL training.}
    \begin{tabular}{| l | l |}
        \hline
        RL Hyperparameter & Value \\
        \hline
        number of steps & $3\,000\,000$ \\
        steps after which learning starts & $100\,000$ \\
        episode time limit (steps) & $100$ \\
        number of parallel environments & $10$ \\
        gradient descent steps per environment step & $1$ \\
        \hline
    \end{tabular}
    \label{tab:rl_hyperparameters}
\end{table}

\fakepar{Evaluation} Figures showing the evaluation success rate over time are slightly smoothed by applying a Gaussian filter with $\sigma = 2.5$ evaluation steps. Evaluations steps are performed every $10\,000$ training steps with 10 evaluation episodes each.

\fakepar{State representation}
In our RL evaluation, we consider different state representations. Table~\ref{tab:rl_state_representations} summarizes these configurations. We distinguish between using SAE-encoded keypoints and using ground truth (GT) points as well as between extending the state with end-effector state measurements (+ robot) or not doing this. Additionally we consider the full simulation state, including target position and cube position, velocity, orientation, and angular velocity. Note that $o$ and $\phi$ denote 3D positions and orientations, respectively, while $\dot{o}$ and $\dot{\phi}$ are their temporal derivatives.
\begin{table}
    \centering
    \caption{Specifications of state representation vector configurations used for RL experiments.}
    \begin{tabular}{| l | l |}
        \hline
        Configuration & State\\
        \hline
        Keypoints & $s^t = z^t$ \\
        Keypoints + robot & $s^t_\mathrm{ext} = (z^t, o_\mathrm{eef}, \dot{o}_\mathrm{eef})$ \\
        GT points & $s^t = x^t$ \\
        GT points + robot & $s^t_\mathrm{ext} = (x^t, o_\mathrm{eef}, \dot{o}_\mathrm{eef})$ \\
        Full Simulation State & $s^t_\mathrm{full} = (o_\mathrm{eef}, \dot{o}_\mathrm{eef}, o_\mathrm{cube}, \dot{o}_\mathrm{cube}, \phi_\mathrm{cube}, \dot{\phi}_\mathrm{cube}, o_\mathrm{target})$ \\
        \hline
    \end{tabular}
    \label{tab:rl_state_representations}
\end{table}

\section{Experimental Results}
In the following, we discuss selected additional results we obtained. These are a comparison of our SAE metric with the reconstruction loss, qualitative results of applying the linear feature transformation, and a closer look at tracking error distributions for the considered SAE architecture modifications.

\begin{figure}[tb]
    \centering
    \subfigure[]{
        \label{fig:basicfp32_loss}
        \includegraphics[width=0.48\textwidth]{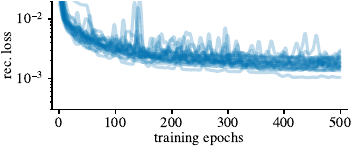}
    }
    \subfigure[]{
        \label{fig:keynet_loss}
        \includegraphics[width=0.48\textwidth]{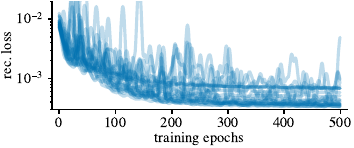}
    }%
    \caption{Reconstruction loss over training epochs for the Basic-kp32 architecture (\ref{fig:basicfp32_loss}) and the KeyNet architecture (Fig. \ref{fig:keynet_loss}).}
     \label{fig:rec_loss}
\end{figure}

\fakepar{Reconstruction loss}
In comparison to tracking errors and tracking capability, we examine the reconstruction loss of two exemplary architectures in Figure~\ref{fig:rec_loss}. Overall, we observe the expected decrease in reconstruction loss over training epochs and a difference in final reconstruction loss between the two architectures. In contrast to the tracking error, however, the reconstruction loss cannot tell us if any of the two architectures effectively tracks the ground truth objects. Neither are any notable jumps present, as they are for the tracking error over time, nor is there any way of differentiating between multiple objects. We, therefore, find the reconstruction loss to be an insufficient indicator for judging an SAE's tracking performance.

\begin{figure}[tb]
    \centering
    \subfigure[Trajectories of the end-effector]{
        \label{fig:trajectories_endeffector}
        \includegraphics[width=\textwidth]{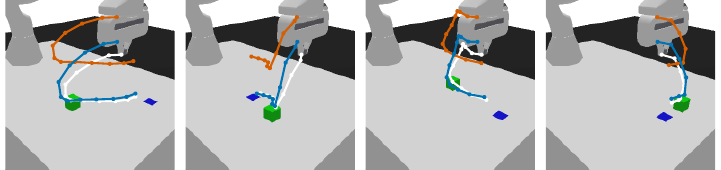}}
    \hspace{2.44mm}
    \subfigure[Trajectories of the cube]{
        \label{fig:trajectories_cube}
        \includegraphics[width=\textwidth]{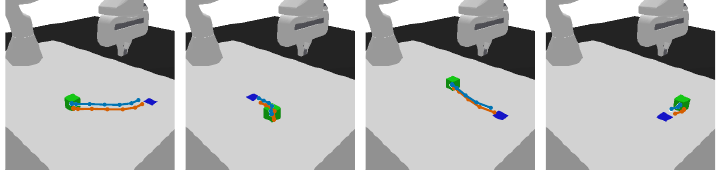}}%
    \caption{Trajectories from sample episodes for end-effector tracking in Figure \ref{fig:trajectories_endeffector} and cube tracking in Figure \ref{fig:trajectories_cube}. Ground truth points (white), untransformed keypoints obtained with KeyNet-vel-std-bg (red), and transformed keypoints (blue) are plotted over time.}
    \label{fig:trajectory_examples}
\end{figure}

\fakepar{Qualitative transformation results}
Figure~\ref{fig:trajectory_examples} shows trajectories from four example episodes for the end-effector and the cube. The trajectory of the best keypoint according to our tracking error formulation is shown in red. As expected, is to be noted that this trajectory can significantly deviate from the white ground truth point trajectory. When taking into account the proposed transformation for tracking evaluation, we obtain the blue trajectory. This trajectory comes much closer to the ground truth point trajectory we evaluate against. For the cube, these to trajectories coincide almost exactly. The larger deviation for the end-effector can be explained by its wider range of movement the transformation for which is approximated linearly and time-invariant.

\begin{figure}[tb]
    \centering
    \includegraphics[width=\textwidth]{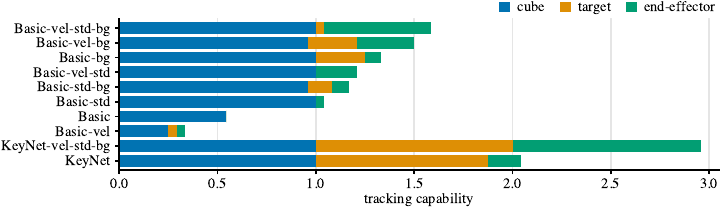}
    \caption{Model ablations: tracking capability $\overline{\mathrm{TC}_k}$ for $k=3$ ground truth objects of the adapted architectures.}
    \label{fig:sae_ablations_tc}
\end{figure}

\fakepar{Modifications and ablations}
Figure~\ref{fig:sae_ablations_tc} shows the tracking capabilities of the Basic architecture with all combinations of modifications and the KeyNet architecture with the best-found combination. The velocity loss (-vel) generally improves the tracking capability for continuously moving objects, in our case, the end-effector. However, it can have a slightly negative effect on tracking of more stationary objects.
Making $\sigma$ trainable (-std) and adding a background bias layer (-bg) almost always yields an improved tracking capability. The combination vel-std-bg yields a consistent improvement in tracking capability. For Basic, it is increased from $\overline{\mathrm{TC}}=0.181$ to $\overline{\mathrm{TC}}=0.528$.
\begin{figure}[tb]
    \centering
    \includegraphics[width=\textwidth]{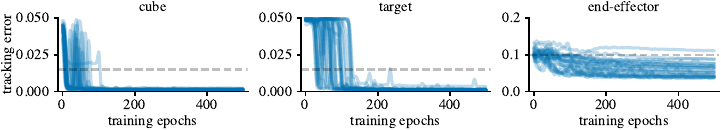}
    \caption{KeyNet-vel-std-bg tracking errors $\overline{\mathrm{TC}_k}$ for $k=3$ ground truth objects over epochs.}
    \label{fig:keynetvelvarbg_tracking_error_time}
\end{figure}
KeyNet already tracks cube and target well and reaches $\overline{\mathrm{TC}}=0.681$. KeyNet-vel-std-bg has a near-perfect mean tracking capability of $\overline{\mathrm{TC}}=0.986$. The biggest change can be seen in end-effector tracking, improving from $\mathrm{TC}_\mathrm{eef}=0.167$ to $0.958$.
This exceptional tracking performance is confirmed by the tracking error. For cube and target the tracking error consistently shows a distinct drop below our threshold (\cf Figure~\ref{fig:keynetvelvarbg_tracking_error_time}). As explained in Section~\ref{sec:sae_eval}, the end-effector usually shows a shallower slope. Still, the tracking error consistently falls below the threshold, indicating that the KeyNet-vel-std-bg architecture learns to track the gripper.

\begin{figure}[tb]
    \centering
    \includegraphics[width=\textwidth]{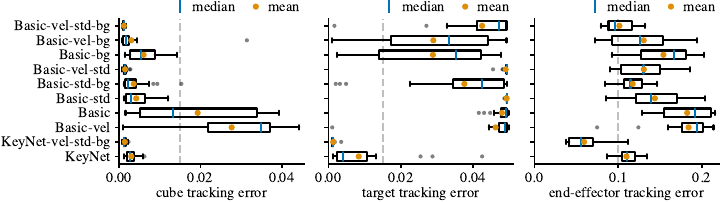}
    \caption{Box plots of tracking errors over 24 runs of the considered architecture modifications.}
    \label{fig:sae_ablations_box}
\end{figure}

Due to its more directly interpretable nature, we primarily examine the tracking capability $\mathrm{TC}$ in our evaluation of architecture modifications. Figure~\ref{fig:sae_ablations_box} additionally shows box plots of the object-wise tracking error distributions over 24 runs for each model. We again observe the effectiveness of the joined modification -vel-std-bg for both the Basic and the KeyNet architecture.
\end{document}